\def\year{2020}\relax
\documentclass[letterpaper]{article} % DO NOT CHANGE THIS
\usepackage{aaai20}  % DO NOT CHANGE THIS
\usepackage{times}  % DO NOT CHANGE THIS
\usepackage{helvet} % DO NOT CHANGE THIS
\usepackage{courier}  % DO NOT CHANGE THIS
\usepackage[hyphens]{url}  % DO NOT CHANGE THIS
\usepackage{graphicx} % DO NOT CHANGE THIS
\usepackage{graphicx}  % DO NOT CHANGE THIS
\usepackage{url}
\usepackage{bm}
\usepackage{multirow}
\usepackage{arydshln}
\usepackage{dsfont}
\usepackage{booktabs}
\usepackage[ruled]{algorithm2e}
\usepackage{amsmath}
\usepackage{multirow}

\usepackage{amssymb,amsmath,amsthm}

\usepackage{xcolor}
\usepackage{subfigure}
\title{Binarized Neural Architecture Search}
\author{Hanlin Chen\textsuperscript{1}, Li'an Zhuo\textsuperscript{1}, Baochang Zhang\textsuperscript{1}\thanks{Baochang Zhang is the corresponding author.}, \\ 
	{\bf \Large Xiawu Zheng\textsuperscript{2},  Jianzhuang Liu\textsuperscript{4}, David Doermann\textsuperscript{3}, Rongrong Ji\textsuperscript{2}}  \\
\textsuperscript{1}Beihang University, 
\textsuperscript{2}Xiamen University,
\textsuperscript{3}University at Buffalo, \\
\textsuperscript{4}Shenzhen Institutes of Advanced Technology, University of Chinese Academy of Sciences\\
\{hlchen, bczhang\}@buaa.edu.cn\\
}
 
\begin{document}

\maketitle

	\begin{abstract}
	Neural architecture search (NAS) can have a significant impact in computer vision by automatically designing optimal neural network architectures for various tasks. A variant, binarized neural architecture search (BNAS), with a search space of binarized convolutions, can produce extremely compressed models.  Unfortunately, this area remains largely unexplored.   BNAS is more challenging than NAS due to the learning inefficiency caused by optimization requirements and the huge architecture  space. To address these issues, we introduce channel sampling and  operation space reduction into a differentiable NAS to significantly reduce the cost of searching. This is accomplished through a performance-based strategy used to abandon less potential operations. Two optimization methods for binarized neural networks are used to validate the effectiveness of our BNAS. Extensive experiments demonstrate that the proposed BNAS achieves a performance comparable to NAS on both CIFAR and ImageNet databases.  An accuracy of $96.53\%$ vs. $97.22\%$ is achieved on the CIFAR-10 dataset, but with a significantly compressed model, and a $40\%$ faster search than the state-of-the-art PC-DARTS.
	\end{abstract}
\section{Introduction}
	
	Neural architecture search (NAS) have attracted great attention with remarkable performance in various deep learning tasks. Impressive results have been shown for reinforcement learning (RL) based methods \cite{zoph2018learning,zoph2016neural}, for example, which train and evaluate more than $20,000$ neural networks across $500$ GPUs over $4$ days. Recent methods like differentiable architecture search (DARTs) reduce the search time by formulating the task in a differentiable manner \cite{liu2018darts}. DARTS relaxes the search space to be continuous, so that the architecture can be optimized with respect to its validation set performance by gradient descent, which provides a fast solution for effective network architecture search. To reduce the redundancy in the network space,  partially-connected DARTs (PC-DARTs) was recently  introduced to perform a more efficient search without compromising the performance of DARTS \cite{xu2019pcdarts}.
	
	Although the network optimized by DARTS or its variants has a smaller model size than  traditional light models,  the searched network   still suffers from an inefficient  inference process due to the complicated  architectures generated by multiple stacked  full-precision convolution operations. Consequently, the adaptation of the searched network to an embedded device is still computationally expensive and inefficient.  Clearly the problem requires further exploration to overcome these challenges.
	
	One way to address these challenges is to  transfer the NAS to a binarized neural architecture search (BNAS), by exploring the advantages of binarized neural networks (BNNs) on memory saving and computational cost reduction \cite{shen2019searching}. Binarized filters have been used in traditional convolutional neural networks (CNNs) to compress deep models \cite{paper09,paper10,paper14,paper15}, showing up to 58-time speedup and 32-time memory saving. In \cite{paper15}, the XNOR network is presented where both the weights and inputs attached to the convolution are approximated with binary values.  This results in an efficient implementation of convolutional operations by reconstructing the unbinarized filters with a single scaling factor. In \cite{gu2019projection}, a projection  convolutional neural network (PCNN) is proposed to realize BNNs based on a simple back propagation algorithm. In our BNAS framework, we re-implement XNOR and PCNN for the effectiveness  validation. We show that the BNNs obtained by BNAS can outperform conventional models by a large margin. It is a significant contribution in the field of BNNs, considering that the performance of conventional BNNs are not yet comparable with their corresponding full-precision models in terms of accuracy. 
	
	\begin{figure*}[htbp!]
		\centering
		\includegraphics[scale=.46]{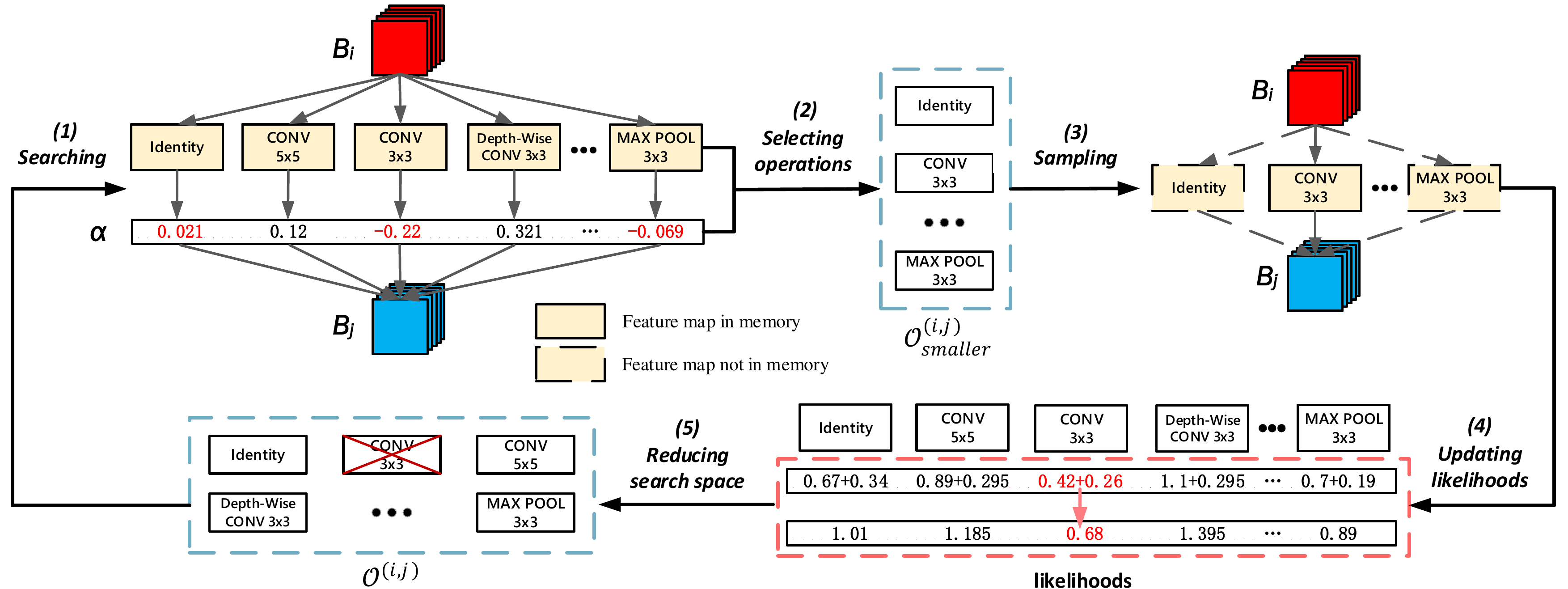}
		\caption{The main steps of our BNAS: (1) Search an architecture based on $\mathcal{O}^{(i,j)}$ using PC-DARTS. (2) Select half the operations with less potential from $\mathcal{O}^{(i,j)}$ for each edge, resulting in $\mathcal{O}^{(i,j)}_{smaller}$. (3) Select an architecture by sampling (without replacement) one operation from $\mathcal{O}^{(i,j)}_{smaller}$ for every edge, and then train the selected architecture. (4) Update the operation selection likelihood $s(o^{(i,j)}_k)$ based on the accuracy obtained from the selected architecture on the validation data. (5) Abandon the operation with the minimal selection likelihood from the search space $\{\mathcal{O}^{(i,j)}\}$ for every edge.}
		\label{fig:performance-based}
	\end{figure*}
	
	The search process of our BNAS consists of two steps. One is the operation potential ordering based on partially-connected DARTs (PC-DARTs) \cite{xu2019pcdarts} which serves as a baseline for our BNAS. It is further sped up with a second operation reduction step guided by a performance-based strategy. In the operation reduction step, we prune one operation at each iteration from one-half of the operations with less potential as calculated by PC-DARTS. As such, the optimization of the two steps becomes faster and faster because the search space is reduced due to the operation pruning. We can take advantage of the differential framework of DARTS where the search and performance evaluation are in the same setting. We also enrich the search strategy of DARTS. Not only is the gradient used to determine which operation is better, but the proposed performance evaluation is included for further reduction of the search space. In this way BNAS is fast and well built. The contributions of our paper include:
	
	\begin{itemize}
		\item
		BNAS is developed based on a new search algorithm which solves the BNNs optimization and architecture search in a unified framework. 
		\item
		The search space is greatly reduced through a performance-based strategy used to abandon  operations with less potential, which improves the search efficiency by $40\%$.
		
		\item
		Extensive experiments demonstrate that the proposed algorithm achieves much better performance than other light models on  CIFAR-10 and ImageNet. 
	\end{itemize}
	
	\section{Related Work}
	Thanks to the rapid development of deep learning, significant gains in performance have been realized in a wide range of computer vision tasks, most of which are manually designed network architectures \cite{krizhevsky2012imagenet,simonyan2014very,he2016deep,huang2017densely}. Recently, the new approach called neural architecture search (NAS) has been attracting increased attention. The goal is to find automatic ways of designing neural architectures to replace conventional hand-crafted ones. Existing NAS approaches need to explore a very large search space and can be roughly divided into three type of approaches: evolution-based, reinforcement-learning-based and one-shot-based.

	In order to implement the architecture search within a short period of time, researchers try to reduce the cost of evaluating each searched candidate. Early efforts include sharing weights between searched and newly generated networks \cite{cai2018efficient}.  Later, this method was generalized into a more elegant framework named one-shot architecture search \cite{brock2017smash,cai2018proxylessnas,liu2018darts,pham2018efficient,xie2018snas,zheng2019multinomial,zheng2019dynamic}. In these approaches, an over-parameterized network or super network covering all candidate operations is trained only once, and the final architecture is obtained by sampling from this super network. For example, Brock et al. \cite{brock2017smash} trained the over-parameterized network using a HyperNet \cite{2016Hypernetworks}, and Pham et al. \cite{pham2018efficient} proposed to share parameters among child models to avoid retraining each candidate from scratch. The paper \cite{liu2017hierarchical} is based on DARTS, which introduces a differentiable framework and thus combines the search and evaluation stages into one. Despite its simplicity, researchers have found some of its drawbacks and proposed a few improved approaches over DARTS \cite{cai2018proxylessnas,xie2018snas,Chen2019SMASH}.
	
	Unlike previous methods, we study BNAS based on efficient operation reduction. We prune one operation at each iteration from one-half of the operations with smaller weights calculated by PC-DARTS, and the search becomes faster and faster in the optimization.

	\section{Binarized Neural Architecture Search}
	
	In this section, we first describe the search space in a general form, where the computation procedure for an architecture (or a cell in it) is represented as a directed acyclic graph. We then review the baseline PC-DARTS \cite{xu2019pcdarts}, which improves the memory efficiency, but is   not enough for BNAS. Finally, an operation sampling and a performance-based search strategy are proposed to effectively reduce the search space. Our BNAS framework is shown in Fig. \ref{fig:performance-based} and additional details of which are described in the rest of this section.

	\subsection{Search Space}
\begin{figure}[htbp]
		\centering
		\subfigure[Cell]{ %[pic1.]
			\includegraphics[scale=.36]{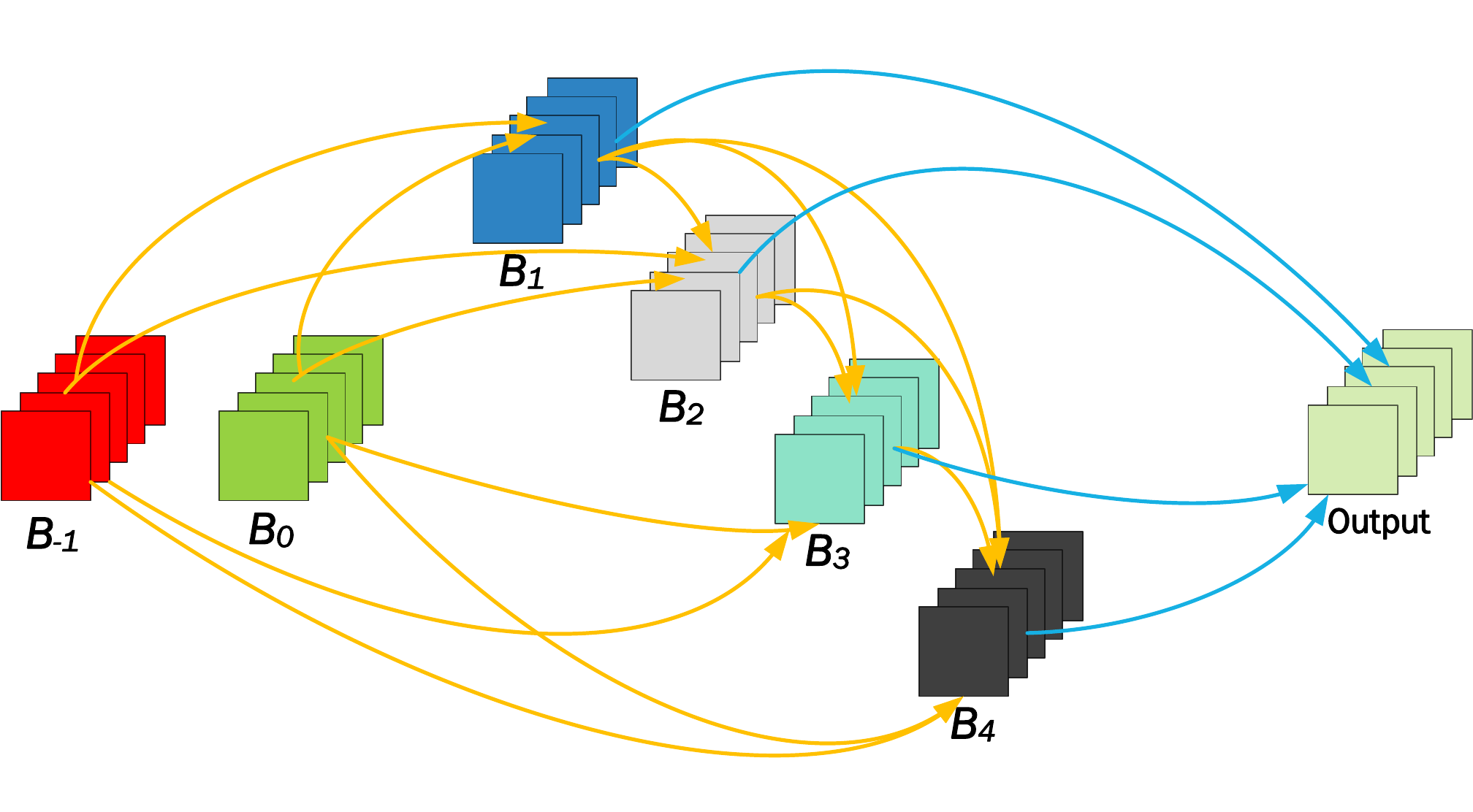}
			\label{fig:cell}
		}
		\quad
		
		\subfigure[Operation Set]{ %[pic2.]
			\includegraphics[scale=.4]{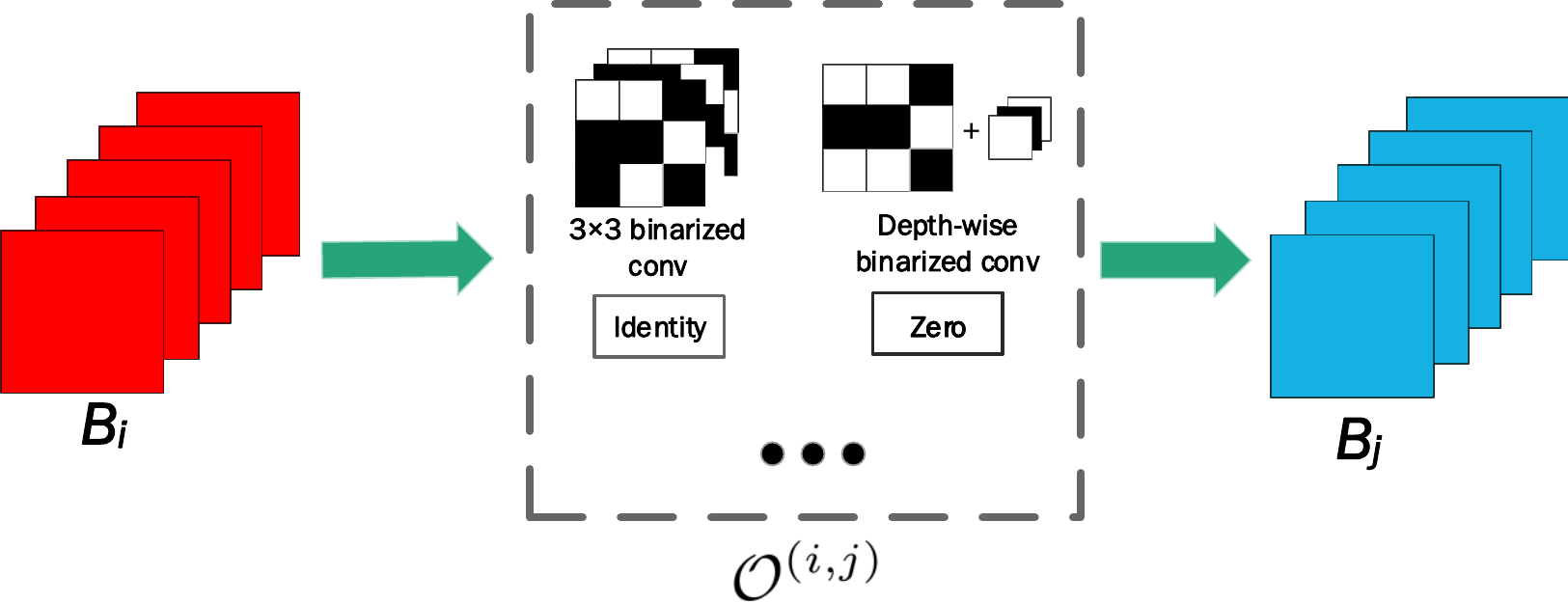}
			\label{fig:operationset}
		}
		\caption{ (a) A cell contains 7 nodes, two input nodes $B_{-1}$ and $B_0$, four intermediate nodes $B_1$, $B_2$, $B_3$, $B_4$ that apply sampled operations on the input nodes and upper nodes, and an output node that concatenates the outputs of the four intermediate nodes. (b) The set of operations $\mathcal{O}^{(i,j)}$ between $B_i$ and $B_j$, including binarized convolutions.}
	\end{figure}

	Following Zoph et al. (2018); Real et al. (2018); Liu et al. (2018a;b), we search for a computation cell as the building block of the final architecture. A network consists of a pre-defined number of cells \cite{zoph2016neural}, which can be either normal cells or reduction cells. Each cell takes the outputs of the two previous cells as input. A cell is a fully-connected directed acyclic graph (DAG) of $M$ nodes, \emph{i.e.}, $\{B_1, B_2, ..., B_M\}$, as illustrated in Fig. \ref{fig:cell}. Each node $B_i$ takes its dependent nodes as input, and generates an output through a sum operation $B_j = \sum_{i<j} o^{(i,j)}(B_i).$ Here each node is a specific tensor (\emph{e.g.,} a feature map in convolutional neural networks) and each directed edge $(i,j)$ between $B_i$ and $B_j$ denotes an operation $o^{(i,j)} (.)$, which is sampled from $\mathcal{O}^{(i,j)}=\{o^{(i,j)}_1, ..., o^{(i,j)}_K\}$. Note that the constraint $i<j$ ensures there are no cycles in a cell. Each cell takes the outputs of two dependent cells as input, and we define the two input nodes of a cell as $B_{-1}$ and $B_{0}$ for simplicity. Following \cite{liu2018darts}, the set of the operations $\mathcal{O}$ consists of $K = 8$ operations.  They include $3\times3$ max pooling, no connection (zero), $3\times3$ average pooling, skip connection (identity), $3\times3$ dilated convolution with rate $2$, $5\times5$ dilated convolution with rate $2$, $3\times3$ depth-wise separable convolution, and $5\times5$ depth-wise separable convolution, as illustrated in Fig. \ref{fig:operationset}. The search space of a cell is constructed by the operations of all the edges, denoted as $\{\mathcal{O}^{(i,j)}\}$.
	
	Unlike conventional convolutions, our BNAS is achieved by transforming all the convolutions in $\mathcal{O}$ to binarized convolutions. We denote the  full-precision and binarized kernels as $X$ and $\hat X$ respectively. A convolution operation in $\mathcal{O}$ is represented as  $B_j = B_i \otimes \hat X $ as shown in Fig. \ref{fig:operationset}, where $\otimes$ denotes convolution. To build BNAS, one key step is how to binarize the kernels from  $X$ to $\hat X$, which can be implemented based on state-of-the-art BNNs, such as XNOR or PCNN. As we know, the optimization of BNNs is more challenging than that of conventional CNNs \cite{gu2019projection,Rastegari2016XNOR}, which adds an additional burden to NAS. To solve it, we introduce channel sampling and  operation space reduction into differentiable NAS to significantly reduce the cost of GPU hours, leading to an efficient BNAS.
	
	\subsection{PC-DARTS}
	%	\subsection{Neural Architecture Search}
	The core idea of PC-DARTS is to take advantage of partial channel connections to improve memory efficiency. Taking the connection from $B_i$ to $B_j$ for example, this involves defining a channel sampling mask $S^{(i,j)}$, which assigns $1$ to selected channels and $0$ to masked ones. The selected channels are sent to a mixed computation of $|\mathcal{O}^{(i,j)}|$ operations, while the masked ones bypass these operations. They are directly copied to the output, which is formulated as:
	
	\begin{equation}
	\label{eq:pc}
	\small
	\begin{aligned}
	&f^{(i,j)}(B_i,S^{(i,j)}) \\
	&= \sum_{o^{i,j}_k\in \mathcal{O}^{(i,j)}} \frac{exp\{\alpha_{o^{(i,j)}_k}\}}{\sum_{o^{(i,j)}_{k^{'}} \in \mathcal{O}^{(i,j)}} exp\{\alpha_{o^{(i,j)}_{k^{'}}}\} } \cdot o^{(i,j)}_k(S^{(i,j)} * B_i) \\
	&+ (1 - S^{(i,j)}) * B_i,
	\end{aligned}
	\end{equation}
	where $S^{(i,j)} * B_i$ and $(1 - S^{(i,j)}) * B_i$ denote the selected and masked channels, respectively, and $\alpha_{o^{(i,j)}_k}$ is the parameter of operation $o^{(i,j)}_k$ between $B_i$ and $B_j$.
	
	PC-DARTS sets the proportion of selected channels to $1/C$ by regarding $C$ as a hyper-parameter. In this case, the computation cost can also be reduced by $C$ times. However, the size of the whole search space is $2 \times K^{|\mathcal{E_M}|}$, where $\mathcal{E_M}$ is the set of possible edges with $M$ intermediate nodes in the fully-connected DAG, and the "$2$" comes from the two types of cells. In our case with $M=4$, together with the two input nodes, the total number of cell structures in the search space is $2 \times 8^{2+3+4+5} = 2 \times 8^{14}$.  This is an extremely large space to search for a binarized neural architectures which need more time than a full-precision NAS. Therefore, efficient optimization strategies for BNAS are required.
	
	\subsection{Sampling for BNAS}
	For BNAS, PC-DARTS is still time and memory consuming because of the large search space, although it is already faster than most of existing NAS methods. We introduce another approach to increasing memory efficiency by reducing the search space $\{\mathcal{O}^{(i,j)}\}$. According to $\{\alpha_{o^{(i,j)}_k}\}$, we can select half the operations with less potential from $\mathcal{O}^{(i,j)}$ for each edge, resulting in $\mathcal{O}^{(i,j)}_{smaller}$. We then sample an operation from $\mathcal{O}^{(i,j)}_{smaller}$ for each edge guided by a performance-based strategy proposed in the next section in order to reduce the search space. We follow the rule of sampling without replacement $K/2$ times. Here sampling without replacement means that after one operation is sampled randomly from $\mathcal{O}^{(i,j)}_{smaller}$, this operation is removed from $\mathcal{O}^{(i,j)}_{smaller}$. For convenience of description, the $K/2$ operations in each edge are transformed to a one-hot indicator vector. In other words we sample only one operation according to the performance-based strategy, which effectively reduces the memory cost compared with PC-DARTS \cite{xu2019pcdarts}.
	
	\subsection{Performance-based Strategy for BNAS} \label{sec:PBS}
	Reinforcement learning is inefficient in the architecture search due to the delayed rewards in network training, i.e., the evaluation of a structure is usually done after the network training converges. On the other hand, we can perform the evaluation of a cell when training the network. Inspired by \cite{ying2019bench}, we use a performance-based strategy to boost the search efficiency by a large margin. Ying \emph{et al.} \cite{ying2019bench} did a series of experiments showing that in the early stage of training, the validation accuracy ranking of different network architectures is not a reliable indicator of the final architecture quality. However, we observe that the experiment results actually suggest a nice property that if an architecture performs badly in the beginning of training, there is little hope that it can be part of the final optimal model. As the training progresses, this observation shows less uncertainty. Based on this observation, we derive a simple yet effective operation abandoning process. During training, along with the increasing epochs, we progressively abandon the worst performing operation in each edge.
	
	To this end, we randomly sample one operation from the $K/2$ operations in $\mathcal{O}^{(i,j)}_{smaller}$ for every edge, then obtain the validation accuracy by training the sampled network for one epoch, and finally assign this accuracy to all the sampled operations. These three steps are performed $K/2$ times by sampling without replacement, leading to each operation having exactly one accuracy for every edge.
	
	We repeat it $T$ times. Thus each operation for every edge has $T$ accuracies $\{y_{k,1}^{(i,j)}, y_{k,2}^{(i,j)}, ..., y_{k,T}^{(i,j)}\}$. Then we define the selection likelihood of the $k$th operation in $\mathcal{O}^{(i,j)}_{smaller}$  for each edge as:
	
	\begin{equation}\label{eq:performance_prob_smaller}
	\small
	s_{smaller}(o^{(i,j)}_k) = \frac{exp\{\bar{y}_k^{(i,j)}\}}{\sum_m exp\{\bar{y}_m^{(i,j)}\}},
	\end{equation}
	where $\bar{y}_k^{(i,j)} = \frac{1}{T}  \sum_t y_{k,t}^{(i,j)}$. And the selection likelihoods of the other operations not in $\mathcal{O}^{(i,j)}_{smaller}$ are defined as:
	
	\begin{equation}\label{eq:performance_prob_larger}
	\small
	\begin{aligned}
	&s_{larger}(o^{(i,j)}_k) \\
	=& \frac{1}{2} (\mathop{\max}\limits_{o^{(i,j)}_k}{\{s_{smaller}(o^{(i,j)}_k)\}} + \frac{1}{\lceil K/2 \rceil} \sum_{o^{(i,j)}_k}{s_{smaller}(o^{(i,j)}_k)}),
	\end{aligned}
	\end{equation}
	where $\lceil K/2 \rceil$ denotes the smallest integer $\geq K/2$. The reason to use it is because $K$ can be an odd integer during iteration  in the proposed Algorithm \ref{alg:MDL}. Eq. \ref{eq:performance_prob_larger} is an estimation for the rest operations using a  value balanced between the maximum and average of $s_{smaller}(o^{(i,j)}_k)$. Then, $s(o^{(i,j)}_k)$ is updated by: %the smallest integer $\geq K/2$
	
	\begin{equation}\label{eq:performance_prob}
	\begin{aligned}
	\smaller
	s(o^{(i,j)}_k) \leftarrow & \frac{1}{2} s(o^{(i,j)}_k) + q_k^{(i,j)}s_{smaller}(o^{(i,j)}_k) + \\
	&  (1 - q_k^{(i,j)})s_{larger}(o^{(i,j)}_k),
	\end{aligned}
	\end{equation}
	where $q_k^{(i,j)}$ is a mask, which is $1$ for the operations in $\mathcal{O}_{smaller}^{(i,j)}$ and $0$ for the others.
	
	Finally, we abandon the operation with the minimal selection likelihood for each edge. Such that the search space size is significantly reduced from $2 \times |\mathcal{O}^{(i,j)}|^{14}$ to $2 \times (|\mathcal{O}^{(i,j)}|-1)^{14}$.  We have:
	
	\begin{equation}\label{eq:min}
		\small
		\mathcal{O}^{(i,j)} \leftarrow \mathcal{O}^{(i,j)} - \{ \mathop{\arg\min}\limits_{o^{(i,j)}_k}{s(o^{(i,j)}_k)} \}.
	\end{equation}
	
	The optimal structure is obtained when there is only one operation left in each edge. Our performance-based search algorithm is presented in Algorithm \ref{alg:MDL}. Note that in line 1, PC-DARTS is performed for $L$ epochs as the warm-up to find an initial architecture, and line 14 is used to update the architecture parameters $\alpha_{o^{(i,j)}_k}$ for all the edges due to the reduction of the search space $\{\mathcal{O}^{(i,j)}\}$.
	
	\begin{algorithm}[h]
		\small
		\caption{Performance-Based Search \label{alg:MDL}}
		\LinesNumbered
		\KwIn{Training data, Validation data, Searching hyper-graph: $\mathcal{G}$, $K=8$, $s(o^{(i,j)}_k)=0$ for all edges;}
		\KwOut{Optimal structure $\alpha$;}
		Search an architecture for $L$ epochs based on $\mathcal{O}^{(i,j)}$ using PC-DARTS; \\
		\While{$(K>1)$}{
			Select $\mathcal{O}^{(i,j)}_{smaller}$ consisting of $\lceil K/2 \rceil$ operations with smallest $\alpha_{o^{(i,j)}_k}$ from $\mathcal{O}^{(i,j)}$ for every edge; \\
			\For{$t= 1,...,T$ \rm epoch}{
				$\mathcal{O}^{'(i,j)}_{smaller} \leftarrow \mathcal{O}^{(i,j)}_{smaller}$; \\
				\For{ $ e=1,..., \lceil K/2 \rceil $ \rm epoch}{
					Select an architecture by sampling (without replacement) one operation from $\mathcal{O}^{'(i,j)}_{smaller}$ for every edge; \\
					Train the selected architecture and get the accuracy on the validation data; \\
					Assign this accuracy to all the sampled operations; \\
				}
			}
			Update $s(o^{(i,j)}_k)$ using Eq. \ref{eq:performance_prob}; \\
			Update the search space \{$\mathcal{O}^{(i,j)}$\} using Eq. \ref{eq:min};\\

			Search the architecture for $V$ epochs based on $\mathcal{O}^{(i,j)}$ using PC-DARTS; \\
			$K = K -1$; \\
		}
	\end{algorithm}
	
	\subsection{Optimization for BNAS}
	In this paper, the binarized kernel weights are computed based on XNOR \cite{Rastegari2016XNOR} or PCNN \cite{gu2019projection}. Both methods are easily implemented in our BNAS framework, and the source code will be publicly available soon.
	
	Binarizing CNNs, to the best of our knowledge, shares the same implementation framework. Without loss of generality, at layer $l$, let $D_i^l$ be the direction of a full-precision kernel $X_i^l$, and $A^l$ be the shared amplitude. For the binarized kernel $\hat X_i^l$ corresponding to $X_i^l$, we have $\hat{X}_i^l = A^l \odot D_i^l$, where $\odot$ denotes the element-wise multiplication between two matrices. We then employ an amplitude loss function to reconstruct the full-precision kernels as:
	
	\begin{equation}
	\small
	L_{A} = \frac{\theta}{2}\sum_{i, l}\|X^l_i-\hat{X}^l_i\|^2 = \frac{\theta}{2}\sum_{i, l}\|X^l_i-  A^l \odot D^l_i\|^2,
	\label{loss_A}
	\end{equation}
	where $D_i^l=sign(X_i^l)$. The element-wise multiplication combines the binarized kernels and the amplitude matrices to approximate the full-precision kernels. The amplitudes $A^l$ are solved in different BNNs, such as \cite{gu2019projection} and \cite{Rastegari2016XNOR}. The complete loss function $L$ for BNAS is defined as:
	\begin{equation}
	L = L_S + L_{A},
	\label{loss_L}
	\end{equation}
	where $L_S$ is the conventional loss function, e.g., cross-entropy.
	
	\section{Experiments}\label{sec:experiment}
	
	In this section, we compare our BNAS with state-of-the-art NAS methods, and also compare the BNNs obtained by our BNAS based on XNOR \cite{Rastegari2016XNOR} and PCNN \cite{gu2019projection}.

	\subsection{Experiment Protocol}
	
	In these experiments, we first search neural architectures on an over-parameterized network on CIFAR-10, and then evaluate the best architecture with a stacked deeper network on the same data set. Then we further perform experiments to search architectures directly on ImageNet. We run the experiment multiple times and find that the resulting architectures only show slight variation in performance, which demonstrates the stability of the proposed method.
	
	\begin{table*}[htbp]
		\small
		\begin{center}
				\begin{tabular}{lcccc}
					\toprule[1pt]
					\multirow{2}{*}{\textbf{Architecture}} & \textbf{Test Error} & \textbf{\# Params} & \textbf{Search Cost} & \textbf{Search} \\
					& \textbf{(\%)} & \textbf{(M)}    & \textbf{(GPU days)} & \textbf{Method} \\
					\hline
					ResNet-18 \cite{he2016deep}  & 3.53  & 11.1 (32 bits) & - & Manual \\
					WRN-22 \cite{zagoruyko2016wide}  & 4.25 & 4.33 (32 bits) & - & Manual  \\
					DenseNet \cite{huang2017densely} & 4.77 & 1.0 (32 bits) & - & Manual \\
					SENet \cite{hu2018squeeze} & 4.05 & 11.2 (32 bits) & - & Manual \\
					\hline
					ResNet-18 (XNOR) & 6.69 & 11.17 (1 bit)  & - & Manual  \\
					ResNet-18 (PCNN) & 5.63 & 11.17 (1 bit)  & - & Manual  \\
					WRN22 (PCNN) \cite{gu2019projection} & 5.69 & 4.29 (1 bit)  & - & Manual \\
					Network in {\cite{mcdonnell2018training}} & 6.13 & 4.30 (1 bit) & - & Manual\\
					\hline
					NASNet-A \cite{zoph2018learning} & 2.65 & 3.3 (32 bits) & 1800 & RL \\
					AmoebaNet-A \cite{real2018regularized} & 3.34 & 3.2 (32 bits) & 3150 & Evolution \\
					PNAS \cite{liu2018progressive} & 3.41 & 3.2 (32 bits) & 225 & SMBO \\
					ENAS \cite{pham2018efficient}  & 2.89 & 4.6 (32 bits) & 0.5 & RL \\
					Path-level NAS \cite{cai2018path} & 3.64 & 3.2 (32 bits) & 8.3 & RL \\
					DARTS(first order) \cite{liu2018darts}  & 2.94 & 3.1 (32 bits) & 1.5 & Gradient-based \\
					DARTS(second order) \cite{liu2018darts} & 2.83 & 3.4 (32 bits) & 4 & Gradient-based \\
					PC-DARTS & 2.78 & 3.5 (32 bits) & 0.15 & Gradient-based \\ % \cite{xu2019pcdarts}
					\hline
					\textbf{BNAS} (full-precision) & 2.84  & 3.3 (32 bits) & 0.08 & Performance-based \\
					\textbf{BNAS} (XNOR) & 5.71 & 2.3 (1 bit) & 0.104 & Performance-based \\
					\textbf{BNAS} (XNOR, larger) & 4.88 & 3.5 (1 bit) & 0.104 & Performance-based \\
					\textbf{BNAS} (PCNN) & 3.94 & 2.6 (1 bit) & 0.09375 & Performance-based \\
					\textbf{BNAS} (PCNN, larger) & 3.47 & 4.6 (1 bit) & 0.09375  & Performance-based \\ 
					\bottomrule[1pt]
				\end{tabular}%}
			\end{center}
			\caption{Test error rates for human-designed full-precision networks, human-designed binarized networks, full-precision networks obtained by NAS, and networks obtained by our BNAS on CIFAR-10. Note that the parameters are $1$ bit in binarized networks, and are 32 bits in full-precision networks. For fair comparison, we select the architectures by NAS with similar parameters ($<$ $5$M). In addition, we also train an optimal architecture in a larger setting, \emph{i.e.,} with more initial channels ($44$ in XNOR or $48$ in PCNN).}
			\label{tab:cifar_results}
		\end{table*}
%		\vspace{2em}

		We use the same datasets and evaluation metrics as existing NAS works \cite{liu2018darts,cai2018path,zoph2018learning,liu2018progressive}. First, most experiments are conducted on CIFAR-10 \cite{krizhevsky2009learning}, which has $50$K training images and $10$K testing images with resolution $32 \times 32$ and from $10$ classes. The color intensities of all images are normalized to $[-1, +1]$. During architecture search, the $50$K training samples of CIFAR-10 is divided into two subsets of equal size, one for training the network weights and the other for finding the architecture hyper-parameters. When reducing the search space, we randomly select $5$K images from the training set as a validation set (used in line 8 of Algorithm \ref{alg:MDL}). To further evaluate the generalization capability, we stack the discovered optimal cells on CIFAR-10 into a deeper network, and then evaluate the classification accuracy on ILSVRC 2012 ImageNet \cite{russakovsky2015imagenet}, which consists of $1,000$ classes with $1.28$M training images and $50$K validation images.
		
		In the search process, we consider a total of $6$ cells in the network, where the reduction cell is inserted in the second and the fourth layers, and the others are normal cells. There are $M=4$ intermediate nodes in each cell. Our experiments follow PC-DARTS. We set the hyper-parameter $C$ in PC-DARTS to $2$ for CIFAR-10 so only $1/2$ features are sampled for each edge. The batch size is set to $128$ during the search of an architecture for $L=5$ epochs based on $\mathcal{O}^{(i,j)}$ (line 1 in Algorithm \ref{alg:MDL}). Note for $5 \le L \le 10$, the larger $L$ has little effect on the final performance, but will cost more search time. We freeze the network hyper-parameters such as $\alpha$, and only allow the network parameters such as filter weights to be tuned in the first $3$ epochs.  Then in the next 2 epochs, we train both the network hyper-parameters and the network parameters. This is to provide an initialization for the network parameters and thus alleviates the drawback of parameterized operations compared with free parameter operations. We also set $T = 3$ (line 4 in Algorithm \ref{alg:MDL}) and $V = 1$ (line 14), so the network is trained less than $60$ epochs, with a larger batch size of $400$ (due to few operation samplings) during reducing the search space. The initial number of channels is $16$. We use SGD with momentum to optimize the network weights, with an initial learning rate of $0.025$ (annealed down to zero following a cosine schedule), a momentum of 0.9, and a weight decay of $5 \times 10^{-4}$. The learning rate for finding the hyper-parameters is set to $0.01$.
		
		After search, in the architecture evaluation step, our experimental setting is similar to \cite{liu2018darts,zoph2018learning,pham2018efficient}. A larger network of $20$ cells ($18$ normal cells and $2$ reduction cells) is trained on CIFAR-10 for $600$ epochs with a batch size of $96$ and an additional regularization cutout \cite{devries2017improved}. The initial number of channels is $36$. We use the SGD optimizer with an initial learning rate of $0.025$ (annealed down to zero following a cosine schedule without restart), a momentum of $0.9$, a weight decay of $3 \times 10^{-4}$ and a gradient clipping at $5$. When stacking the cells to evaluate on ImageNet, the evaluation stage follows that of DARTS, which starts with three convolution layers of stride $2$ to reduce the input image resolution from $224 \times 224$ to $28 \times 28$. $14$ cells ($12$ normal cells and $2$ reduction cells) are stacked after these three layers, with the initial channel number being $64$. The network is trained from scratch for $250$ epochs using a batch size of $512$. We use the SGD optimizer with a momentum of $0.9$, an initial learning rate of $0.05$ (decayed down to zero following a cosine schedule), and a weight decay of $3 \times 10^{-5}$. Additional enhancements are adopted including label smoothing and an auxiliary loss tower during training. All the experiments and models are implemented in PyTorch \cite{paszke2017automatic}.

		\subsection{Results on CIFAR-10}
		We compare our method with both manually designed networks and networks searched by NAS. The manually designed networks include ResNet \cite{he2016deep}, Wide ResNet (WRN) \cite{zagoruyko2016wide}, DenseNet \cite{huang2017densely} and SENet \cite{hu2018squeeze}. For the networks obtained by NAS, we classify them according to different search methods, such as RL (NASNet \cite{zoph2018learning}, ENAS \cite{pham2018efficient}, and Path-level NAS \cite{cai2018path}), evolutional algorithms (AmoebaNet \cite{real2018regularized}), Sequential Model Based Optimization (SMBO) (PNAS \cite{liu2018progressive}), and gradient-based methods (DARTS \cite{liu2018darts} and PC-DARTS \cite{xu2019pcdarts}).
		
		The results for different architectures on CIFAR-10 are summarized in Tab. \ref{tab:cifar_results}. Using BNAS, we search for two binarized networks based on XNOR \cite{Rastegari2016XNOR} and PCNN \cite{gu2019projection}. In addition, we also train a larger XNOR variant with $44$ initial channels and a larger PCNN variant with $48$ initial channels. We can see that the test errors of the binarized networks obtained by our BNAS are comparable to or smaller than those of the full-precision human designed networks, and are significantly smaller than those of the other binarized networks.

		\begin{table*}[htbp]
			\small
			\begin{center}
				\setlength{\tabcolsep}{3mm}{
					\begin{tabular}{lccccc}
						\toprule
						\multirow{2}{*}{\textbf{Architecture}} & \multicolumn{2}{c}{\textbf{Accuracy (\%)}} & \textbf{Params} & \textbf{Search Cost} & \textbf{Search} \\ \cline{2-3}
						& \textbf{Top1} & \textbf{Top5} & \textbf{(M)} & \textbf{(GPU days)} & \textbf{Method} \\ 
						\hline
						ResNet-18 \cite{gu2019projection} & 69.3 & 89.2 & 11.17 (32 bits) & - & Manual \\
						MobileNetV1 \cite{howard2017mobilenets} & 70.6 & 89.5 & 4.2 (32 bits) & - & Manual \\
						\hline
						ResNet-18 (PCNN) \cite{gu2019projection} & 63.5 & 85.1 & 11.17 (1 bit) & - & Manual \\
						\hline
						NASNet-A \cite{zoph2018learning} & 74.0  & 91.6 & 5.3 (32 bits) & 1800 & RL \\
						AmoebaNet-A \cite{real2018regularized} & 74.5 & 92.0 & 5.1 (32 bits) & 3150 & Evolution \\
						AmoebaNet-C \cite{real2018regularized} & 75.7 & 92.4 & 6.4 (32 bits) & 3150 & Evolution \\
						PNAS \cite{liu2018progressive} & 74.2 & 91.9 & 5.1 (32 bits) & 225 & SMBO \\
						DARTS \cite{liu2018darts} & 73.1 & 91.0 & 4.9 (32 bits) & 4 & Gradient-based \\
						PC-DARTS \cite{xu2019pcdarts} & 75.8 & 92.7 & 5.3 (32 bits) & 3.8 & Gradient-based \\
						\hline
						BNAS (PCNN) & 71.3 & 90.3 & 6.2 (1 bit)  & \textbf{2.6} & Performance-based \\
						\bottomrule
					\end{tabular}}
				\end{center}
				\caption{Comparison with the state-of-the-art image classification methods on ImageNet. BNAS and PC-DARTS are obtained directly by NAS and BNAS on ImageNet, others are searched on CIFAR-10 and then directly transferred to ImageNet.}
				\label{tab:imagenet_results}
			\end{table*}
						
			Compared with the full-precision networks obtained by other NAS methods, the binarized networks by our BNAS have comparable test errors but with much more compressed models. Note that the numbers of parameters of all these searched networks are less than 5M, but the binarized networks only need $1$ bit to save one parameter, while the full-precision networks need $32$ bits. In terms of search efficiency, compared with the previous fastest PC-DARTS, our BNAS is $40\%$ faster (tested on our platform (NVIDIA GTX TITAN Xp). We attribute our superior results to the proposed way of solving the problem with the novel scheme of search space reduction.

\begin{figure}[h]
					\centering
					\subfigure[Normal Cell]{
						\includegraphics[scale=.27]{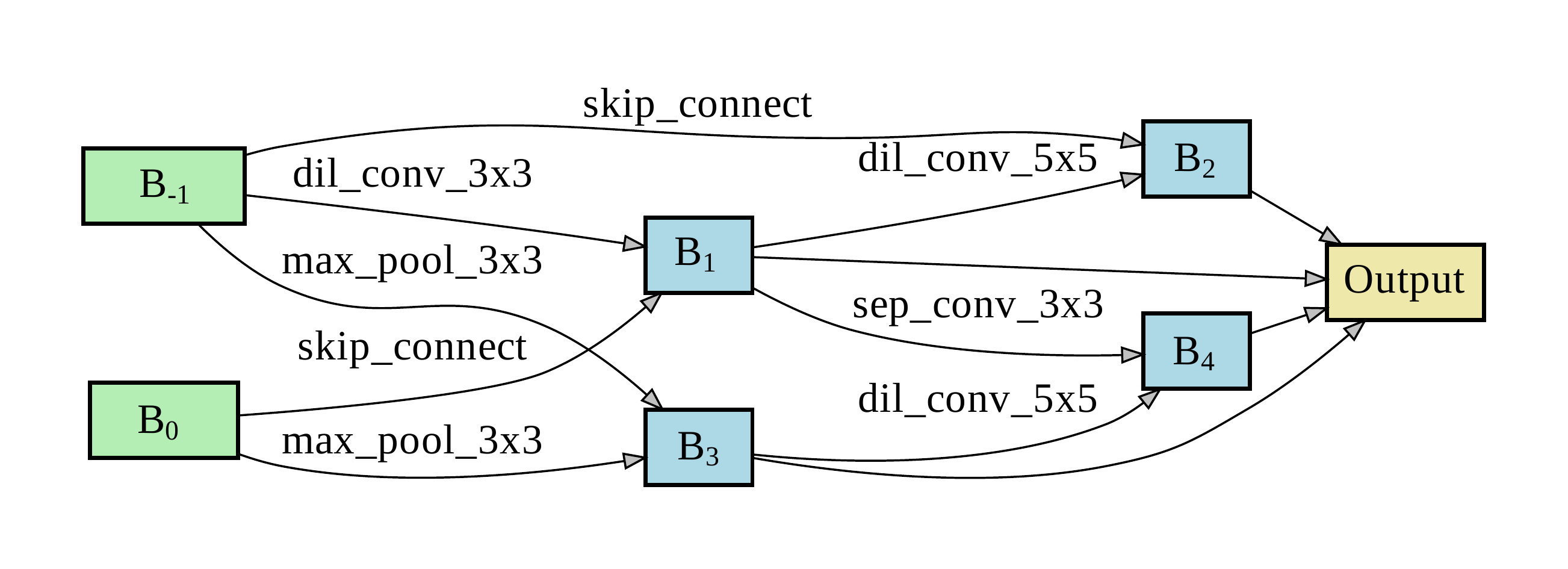}
					}
					\quad
					\subfigure[Reduction Cell]{
						\includegraphics[scale=.18]{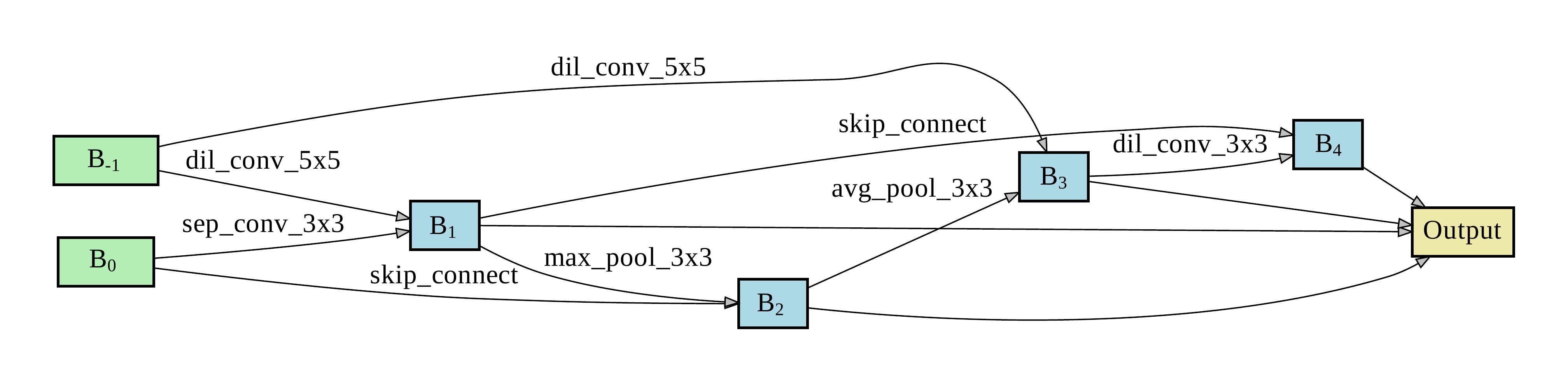}
					}
					\caption{Detailed structures of the best cells discovered on CIFAR-10 using BNAS based on XNOR. In the normal cell, the stride of the operations on $2$ input nodes is 1, and in the reduction cell, the stride is 2.}
					\label{fig:xnor}
				\end{figure}

\begin{figure}[h]
				\centering
				\subfigure[Normal Cell]{
					\includegraphics[scale=.27]{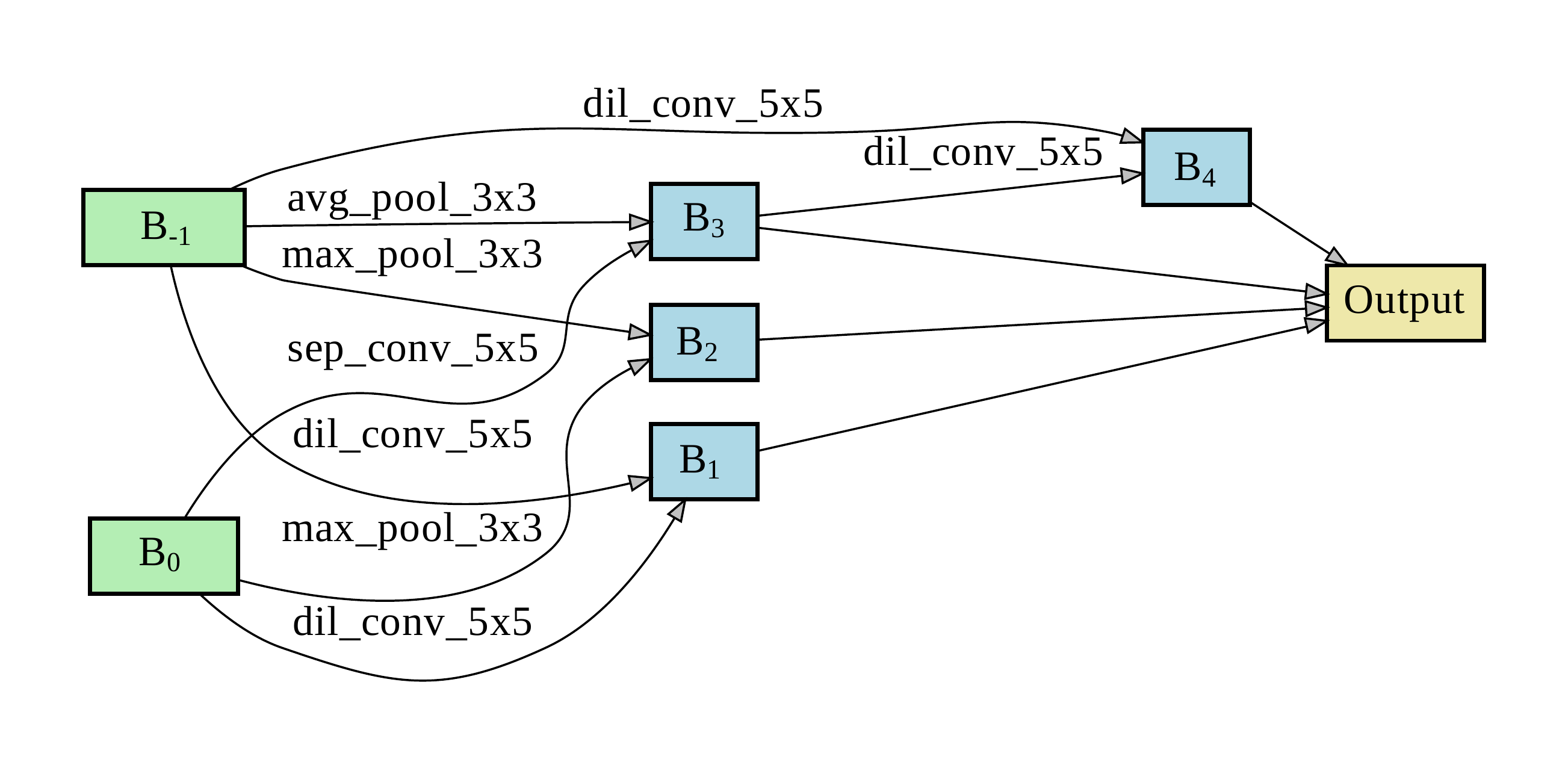}
				}
				\quad
				\subfigure[Reduction Cell]{
					\includegraphics[scale=.27]{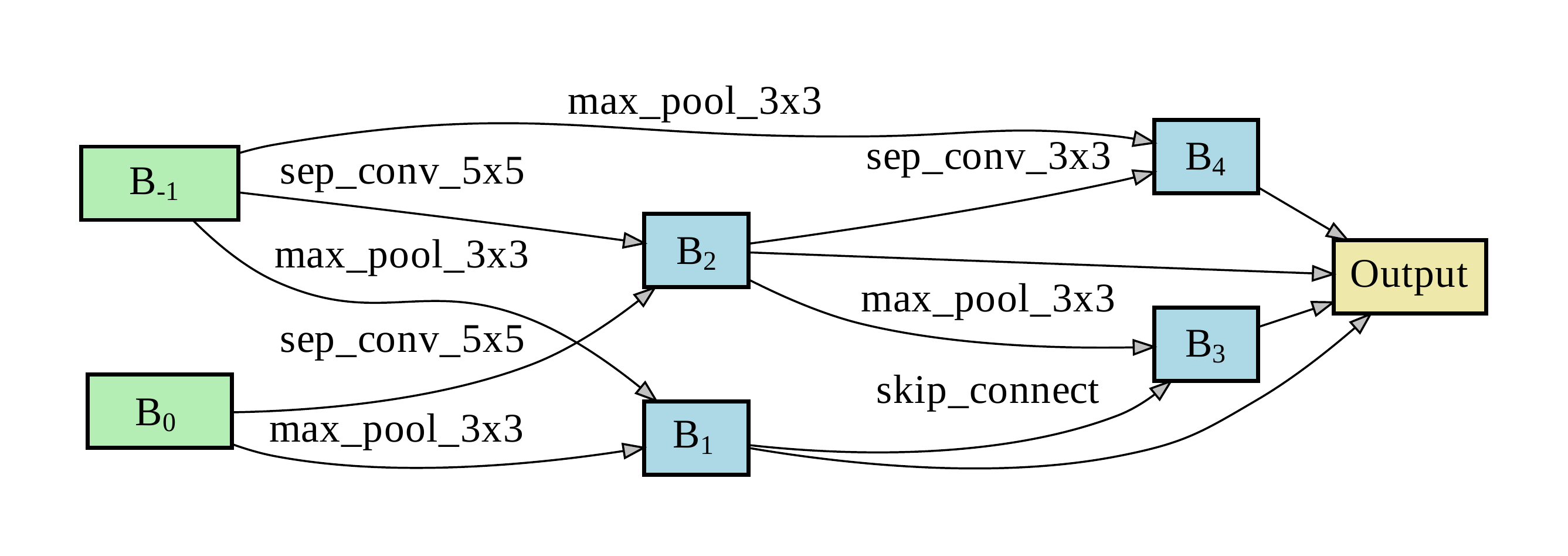}
				}
				\caption{Detailed structures of the best cells discovered on CIFAR-10 using BNAS based on PCNN. In the normal cell, the stride of the operations on $2$ input nodes is 1, and in the reduction cell, the stride is 2.}
				\label{fig:mcn}
			\end{figure}

			Our BNAS method can also be used to search full-precision networks. In Tab. \ref{tab:cifar_results}, BNAS (full-precision) and PC-DARTS perform equally well, but BNAS is $47\%$ faster. Both the binarized methods XNOR and PCNN in our BNAS perform well, which shows the generalization of BNAS. Fig. \ref{fig:xnor} and Fig. \ref{fig:mcn} show the best cells searched by BNAS based on XNOR and PCNN, respectively.
			
			We also use PC-DARTS to perform a binarized architecture search based on PCNN on CIFAR10, resulting in a network denoted as PC-DARTS (PCNN). Compared with PC-DARTS (PCNN), BNAS (PCNN) achieves a better performance ($95.12$\% vs. $96.06$\% in test accuracy) with less search time ($0.18$ vs. $0.09375$ GPU days). The reason for this may be because the performance based strategy can help find bet

			\noindent ter operations for recognition.
			
			\subsection{Results on ImageNet}
			We further compare the state-of-the-art image classification methods on ImageNet. All the searched networks are obtained directly by NAS and BNAS on ImageNet by stacking the cells. Our binarized network is based on PCNNs. From the results in Tab. \ref{tab:imagenet_results}, we have the following observations: (1) BNAS (PCNN) performs better than human-designed binarized networks { (71.3\% vs. 63.5\%)} and has far fewer parameters { (6.1M vs. 11.17M)}. (2) BNAS (PCNN) has a performance similar to the human-designed full-precision networks { (71.3\% vs. 70.6\%)},  with a much more highly compressed model. (3) Compared with the full-precision networks obtained by other NAS methods, BNAS (PCNN) has little performance drop, but is fastest in terms of search efficiency { (0.09375 vs. 0.15 GPU days)} and is a much more highly compressed model due to the binarization of the network. The above results show the excellent transferability of our BNAS method.

			\section{Conclusion}
			
			In this paper, we have proposed BNAS, the first binarized neural architecture search algorithm, which effectively reduces the search time by pruning the search space in early training stages. It is faster than the previous most efficient search method PC-DARTS. The binarized networks searched by BNAS can achieve excellent accuracies on CIFAR-10 and ImageNet. They perform comparable to the full-precision networks obtained by other NAS methods, but with much compressed models.

	\section{Acknowledgements}
	The work was supported in part by National Natural Science Foundation of China under Grants 61672079, 61473086, 61773117, 614730867. This work is supported by Shenzhen Science and Technology Program KQTD2016112515134654. Baochang Zhang is also with Shenzhen Academy of Aerospace Technology, Shenzhen 100083, China.

		\bibliographystyle{aaai}
		\bibliography{1690-aaai}

\end{document}